\documentclass[letterpaper]{article} 
\usepackage{aaai23}  
\usepackage{times}  
\usepackage{helvet}  
\usepackage{courier}  
\usepackage[hyphens]{url}  
\usepackage{graphicx} 
\usepackage{natbib}  
\usepackage{caption} 
\usepackage{algorithm}
\usepackage{algorithmic}

\usepackage{newfloat}
\usepackage{listings}
\DeclareCaptionStyle{ruled}{labelfont=normalfont,labelsep=colon,strut=off} 
\floatstyle{ruled}
\newfloat{listing}{tb}{lst}{}
\floatname{listing}{Listing}
\usepackage{graphicx}
\usepackage{amsmath}
\usepackage{booktabs}
\usepackage{microtype}
\usepackage{algorithm}
\usepackage{algorithmic}
\usepackage{amsfonts}
\usepackage{array}
\usepackage{multirow}
\usepackage{color}

\title{Generalizing Math Word Problem Solvers via Solution Diversification}
\author{
    Zhenwen Liang\textsuperscript{\rm 1},
    Jipeng Zhang\textsuperscript{\rm 2},
    Lei Wang\textsuperscript{\rm 3},
    Yan Wang\textsuperscript{\rm 4},
    Jie Shao\textsuperscript{\rm 5},
    Xiangliang Zhang\textsuperscript{\rm 1}\thanks{Corresponding Author}
}

\affiliations {
    \textsuperscript{\rm 1} University of Notre Dame
    \textsuperscript{\rm 2}  Hong Kong University of Science and Technology
    \textsuperscript{\rm 3} Singapore Management University \\
    \textsuperscript{\rm 4} Tencent AI Lab
    \textsuperscript{\rm 5} University of Electronic Science and Technology of China \\
    zliang6@nd.edu, jzhanggr@conect.ust.hk, lei.wang.2019@phdcs.smu.edu.sg, yanwang.branden@gmail.com, shaojie@uestc.edu.cn, xzhang33@nd.edu
}

\usepackage{bibentry}
\begin{document}

\maketitle

\begin{abstract}
Current math word problem (MWP) solvers are usually Seq2Seq models  trained by the (\emph{one-problem; one-solution}) pairs, each of which is made of a \emph{problem} description and a \emph{solution} showing reasoning flow to get the correct answer. However, one MWP \emph{problem} naturally has \emph{multiple solution} equations. The training of an MWP solver with (\emph{one-problem; one-solution}) pairs excludes other correct solutions, and thus limits the generalizability of the MWP solver. One feasible solution to this limitation is to augment multiple solutions to a given problem. However, it is difficult to collect diverse and accurate augment solutions through human efforts.
In this paper, we design a new training framework for an MWP solver by introducing a \emph{solution buffer} and a \emph{solution discriminator}. The \emph{buffer} includes solutions generated by an MWP solver to encourage  the training data diversity.  The \emph{discriminator} controls the quality of buffered solutions to participate in training. Our framework is flexibly applicable to a wide setting of fully, semi-weakly and weakly supervised training for all Seq2Seq MWP solvers. We conduct extensive experiments on a benchmark dataset Math23k and a new dataset named Weak12k, and show that our framework improves the performance of various MWP solvers under different settings by generating correct and diverse solutions.
\end{abstract}

\section{Introduction}
Automatic math word problem (MWP) solving has attracted the interest of researchers for a long time. Most state-of-the-art MWP solvers \cite{wang2017deep,wang2019template,liu2019tree,xie2019goal,zhang2020graph,shen2020solving,hong2021weakly,liang2021mwp,hu2022solving} are Seq2Seq models, which use an encoder to get latent representations for the problem and a decoder to generate symbolic solutions. The Seq2Seq models are typically trained by  the (\emph{one-problem; one-solution}) pairs, each of which is made of a \emph{problem} of an MWP description and a \emph{solution} of a corresponding ground truth equation. However, MWPs typically 
have multiple reasonable solutions to reach the final answer.  As shown in Table \ref{tab:data_demo}, there exists an alternative solution that gives the same answer as the given ground truth. 

\begin{table}[t]
\renewcommand\arraystretch{1.1}
\centering
\begin{tabular}{|m{2.0cm}<{\centering}|m{4.4cm}<{\centering}|}
\hline
Problem Description: & There are 40 students taking Chinese and math exams, 25 students passed the Chinese exam, 20 students passed the math exam, 10 students failed both exams. How many students pass both exams? \\ 
\hline
Ground Truth Solution:  &  25+20-(40-10) = 15    \\
\hline
Alternative Solution 1: &  25+20-40+10 = 15 \\
\hline
Alternative Solution 2: &  10+25+20-40 = 15 \\
\hline
Spurious Solution: & 25-10=15 \\
\hline
\end{tabular}
\caption{An MWP example with a ground truth solution and an alternative solution that reaches the same correct answer. It also has a spurious solution, which reaches the same correct answer but makes no sense. 
Our target is to improve the generalizability of an MWP solver by considering the multiple solutions to one problem and distinguishing the correct solutions from the spurious solutions. 
}
\label{tab:data_demo}
\end{table}

Training an MWP solver with (\emph{one-problem; one-solution}) pairs excludes other correct solutions, and thus limits the generalizability of the MWP solver. 
An intuitively better idea is to incorporate multiple solutions instead of a single solution to specific MWP in training, i.e., by data augmentation. However, it is difficult to acquire diverse and accurate augment solutions. If randomly generating solutions and then fixing the wrong ones as done in \cite{hong2021weakly}, spurious solutions like ``25-10=15''  shown in Table \ref{tab:data_demo} cannot be identified. They reach the right value answer with the quantities that appeared in the MWP description but make no sense. The inclusion of them in training is harmful to the MWP solver and eventually lowers the performance.
 
We are thus motivated to inject solution diversity into the MWP solver training by meanwhile making quality control on the training instances. One possible way is to label problems manually with diverse solutions. However, this is too time-consuming and costly. Another baseline is based on mathematical transformation to generate all possible variants of ground truth. The critical disadvantage of this transformation is that it generates too many solutions and makes the training target much more ambiguous, leading to poor performance. For example, we can generate 24 potential solutions for $A+B+C+D$ by only using the commutative law. Not to mention the scenario when we consider associative and distributive law if multiplications and parentheses are included. Therefore, we only encourage the potential solutions from the MWP solver itself. 
In fact, to improve the generalizability of MWP solvers, we should make a trade-off between the predictability and diversity of augmented solutions. Therefore, we propose a new training framework for an MWP solver by introducing a \emph{solution buffer} and a \emph{solution discriminator}. The \emph{buffer} is maintained to encourage  the training data diversity by including multiple  solutions  generated by an MWP solver for one given problem.  The \emph{discriminator} controls the quality of buffered solutions to participate in training by calculating their qualification scores.  The solutions with higher qualification scores are allowed to contribute more in future training.

The whole life-cycle of our designed training framework  goes as follows: \underline{i)} update the encoder and decoder parameters of the MWP solver by samples in the solution buffer in a probabilistic way, i.e., higher-quality solutions are more engaged; \underline{ii)} generate new solutions to the buffer by the updated solver; and \underline{iii)} training a diversity-aware discriminator to evaluate the quality of MWP solutions and go to step \underline{i)}. In the iterative training process, a better-trained solver generates higher-quality novel solutions. More diverse solutions train a stronger encoder and decoder, which give a more accurate probabilistic estimation of the solution quality. 
Since the buffered solutions participate in the training of encoder and decoder with different levels of probabilistic weights, our proposed framework well controls the augmentation quality while injecting diversity in training, regardless of the type of Seq2Seq model implemented in the solver.

We conduct extensive experiments to evaluate the effectiveness of our proposed training framework on two datasets. Math23k, the most commonly used dataset for MWP solving, is used to evaluate our training method in the fully and weakly supervised setting. We also curate a novel and large dataset named Weak12k, which has only problems annotated with answer values instead of solutions. By combining Math23k and Weak12k, we
train the model in semi-weakly supervised setting, and then 
evaluate the performance on their testing sets separately. The experimental results show that our proposed training method can generally boost the solving accuracy for different backbone solvers in various experimental settings. The solution diversity evaluation and case studies are strong proofs to show that our method can generate multiple solutions and refine the quality of the training target.

\section{Related Work}
\subsection{Math Word Problem Solving}

After the wide usage of traditional statistical algorithms \cite{hosseini2014learning,mitra2016learning} and semantic parsing methods \cite{shi2015automatically,huang2017learning,liang2018meaning,zou2019text2math} in MWP solving, deep learning methods become dominant. \cite{wang2017deep} first proposed to apply sequence-to-sequence (Seq2Seq) framework to solve MWP and achieved better performance compared with previous methods. Most following works focused on the generation module. \cite{wang2019template} proposed a two-stage decoding method to decompose goals into two parts. \cite{liu2019tree,xie2019goal} proposed to use tree structure decoder. \cite{DBLP:conf/naacl/Chiang2019} introduced a stack-related decoder. Multiple decoder architectures~\cite{ijcai2020-555,shen2020solving} were also introduced to improve generation results. On the other hand, a couple of works \cite{li2019modeling,wang2018translating,lin2021hms} focused on improving the encoding framework. \cite{zhang2020graph,shen2020solving,cao2021bottom} chose to model quantity information with a sequential combination of RNN and GNN encoder. Besides the model architectures, there are also other interesting explorations, such as knowledge distillation \cite{ijcai2020-555}, situation model \cite{hong2021smart}, syntax-semantics model\cite{lyu2021solving}, ,auxiliary training tasks \cite{qin2021neural,piekos2021measuring,ijcai2021-485}, explicit value encoding \cite{wu2021math} and transfer learning\cite{alghamdi2022armath}. Recently, pre-trained language models  \cite{yu2021improving,huang2021recall,DBLP:conf/emnlp/ShenYLSJ0021,li2021seeking,liang2021mwp,lan2022mwptoolkit,liang2022analogical} are widely applied to encode MWPs and become the strongest baselines in terms of MWP solving accuracy. in There are some other works \cite{ran2019numnet,andor2019giving,chen2020neural} considering the weak supervision environment in numerical understanding, however, the solution diversity in MWP is unique and under-explored.

The prior work LBF \cite{hong2021weakly} presents a weakly-supervised MWP solver by considering the diversity of solutions. Our work differs completely from it on both the training framework and the  MWP setting. Our proposed training framework manages the augmented multiple solutions of one MWP with their  probabilistic qualification scores. The harmful influence of  spurious solutions existing in LBF is  alleviated, while the diversity is injected.  
Our training framework is also flexibly usable in  different settings, i.e. full, semi-weak and weak supervision. This is also the first work that considers both the quality and diversity of solutions.

\section{Methodology}

\begin{algorithm}[t]
\small
\caption{Weak Data Augmentation}
\label{alg:algorithm}
\textbf{Input}: Answer $A$ and quantity set $V_{num}$ of problem $W$\\
\textbf{Parameter}: Constant values $V_{con}$ and operators $V_{op}$ of $W$, maximum iteration number $MAX$ \\
\textbf{Output}: Equation $S = \{s_1,s_2,...,s_n\}$ for $W$ 

\begin{algorithmic}[1] 
\STATE Let $R_1 = V_{num} \cup V_{con}, R_2 = \{\}, R_3 = V_{num} \cup V_{con}$ 
\STATE Let $iter = 0$ denote the current iteration number

\WHILE{TRUE}
\FOR{$i \in R_1$}
\FOR{$j \in R_3$}
\FOR{$op \in V_{op}$}
\IF {$iter = MAX$}
\STATE \textbf{Return} failure
\ENDIF
\STATE $S \gets$ connect $i$ and $j$ with operator $op$
\IF {$S$ leads to $A$}
\STATE \textbf{Return} $S$
\ELSE
\STATE Insert $S$ into $R_2$ 
\ENDIF
\STATE $iter = iter + 1$
\ENDFOR
\ENDFOR
\ENDFOR
\STATE $R_3 = R_2 \cup R_3, R_1 = R_2, R_2 = \{\}$
\ENDWHILE
\end{algorithmic}
\end{algorithm}

\subsection{Problem Formulation}
We denote an MWP description as $W = \{w_1,w_2,...,w_m\}$ with length $m$ and its equation-shaped solution as $S = \{s_1,s_2,...,s_n\}$ with length $n$. We let $A$ be the final answer value that can be calculated from the equation $S$. Next, we define a vocabulary for solution $S$ as $V = \{V_{op},V_{num},V_{con}\}$, where $V_{op} = \{+, -, \times, \div, \wedge \}$ contains the operators and $V_{con} = \{1, \pi\}$ contains the constant values that could be used in the solution. The sets $V_{num}$ are created by number mapping \cite{wang2017deep} and have different lengths for different problems, which contain all the numerical quantities appeared in the problem description. 

The objective of an MWP solver is to translate $W$ into $S$, reaching the answer  $A$. An MWP solver can be trained in different settings, by having problem-equation ($W$-$S$) pairs in a \underline{fully supervised} setting, and by having problem-value ($W$-$A$) pairs  in a \underline{weakly supervised} setting. When having a mixture of  problem-equation and problem-value pairs, we investigate a \underline{semi-weakly supervised} setting.
\textcolor{black}{In all these settings, to leverage diverse augmented solutions, we design a buffer to   store and evaluate the qualification scores of these augmented solutions. The buffer is initialized by the provided equation $S$ in the fully supervised or semi-weakly supervised setting. However, in weak supervision with only problem-value ($W$-$A$) pairs, we  initialize the buffer starting from $A$ by a proposed weak data augmentation method.}

\subsection{Weak Data Augmentation}
The weak data augmentation (WDA) method is to search potential equations based on the given value $A$ for problem $W$. With the quantities   $V_{num} \cup V_{con}$ and operators  $V_{op}$ in $W$, the augmentation process can be formulated as an algorithm that receives $V_{num}$ and $A$ as inputs and generates an equation $S$, with $V_{con}$ and $V_{op}$ as parameters that can be adjusted to the dataset. Although some generated solutions might be spurious, our discriminator
could assign small weights to them and alleviate the side effect. In general, our Weak Data Augmentation (WDA) is an equation-orientated neuro-symbolic search algorithm. Details of the proposed WDA can be found in the Algorithm \ref{alg:algorithm}.

The augmentation process is based on  three sets $R_1, R_2$ and $R_3$, where $R_1$ contains all the newly reached solutions in the last round, $R_2$ is a temporary set to store the new solutions in the current loop, and $R_3$ has all the potential solutions founded from the beginning.  Firstly, $R_2$ is initialized as empty, while $R_1$ and $R_3$ are initialized as the combination of quantity set $V_{num}$ and constant values $V_{con}$. 
Then we perform two for-loops to select two equations from $R_1$ and $R_3$ and connect them with an operator in $V_{op}$ to construct a new equation $S$. We iterate $i$ and $j$ on $R_1$ and $R_3$ to avoid duplicated generations. If the new equation $S$ leads to the correct answer, we return the result and terminate the algorithm. Otherwise, $S$ will be inserted into $R_2$ for future usage (line 20 and 21). To avoid the endless trial  without a successful $S$, we   define a maximum iteration number $MAX$ to limit the time consumption on finding $S$, which is set as $50000$ empirically. A failure is returned when no valid equation for a given problem can be found within $MAX$ iterations. 

To reduce the chance of generating spurious solutions and also reduce the searching time, we also formulate several hand-crafted rules. Firstly, we avoid the equations of $a \div a$ and $a - a$ ($a$ represents a random number). Because 1 is given as a constant number and 0 is meaningless for MWPs. Secondly, we discard all solutions with only numbers in $V_{con}$, without any quantity in $V_{num}$. This is because MWPs are supposed to be solved by the quantities $V_{num}$ in the problem description, assisted by $V_{con}$ like 1 and $\pi$, not just by $V_{con}$. Thirdly, there have to be multiplication operators in the solution when constant $\pi$ appears. 

\subsection{Model Architecture}
\begin{figure*}
\centering 
\includegraphics[width=0.95\textwidth]{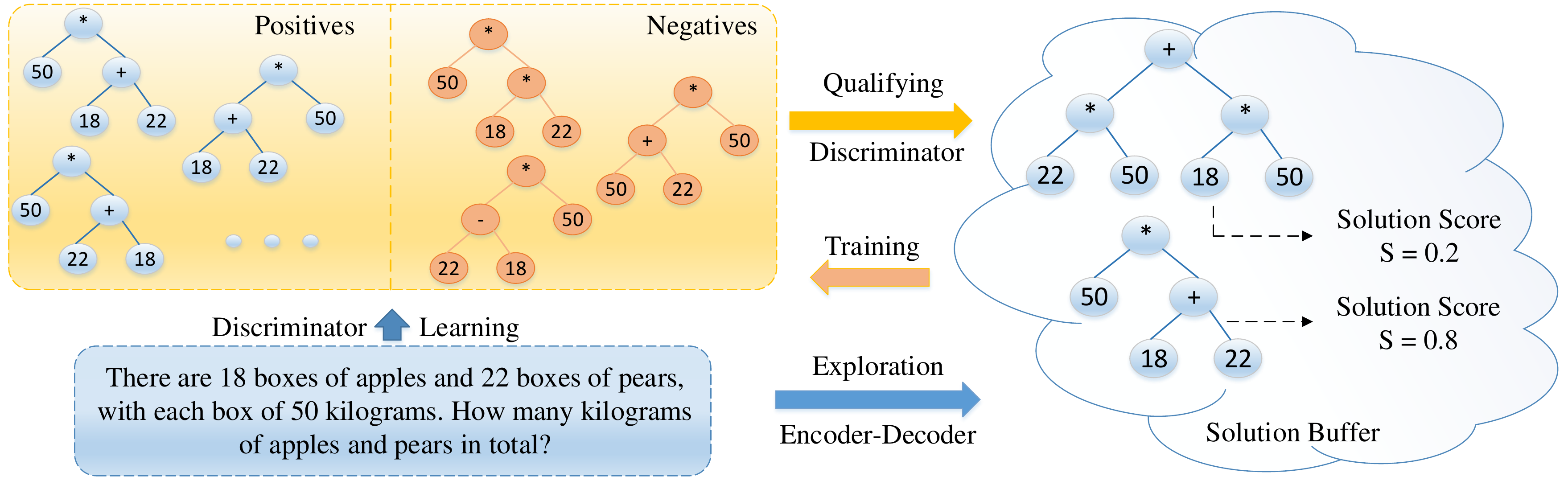} 
\caption{An overview of our proposed framework. The encoder-decoder solver is trained to generate multiple solutions to enlarge the solution buffer. Next, a discriminator is trained with positive samples and negative samples to evaluate the qualified  solution. The solver is then trained again with all potential solutions in the buffer with different weights which are given by the encoder-decoder solver and the discriminator. 
} 
\label{fig:network} 
\end{figure*}

Our proposed framework is shown in Figure \ref{fig:network}. It is a general framework that can host any kind of encoder-decoder solvers. 
\paragraph{Encoder.}
Since GTS \cite{xie2019goal} and MWP-BERT \cite{liang2021mwp} have been recognized as the most representative MWP solvers 
with RNN backbone and pre-trained model backbone, respectively. They are also open-sourced and easy to reproduce the results. Therefore, they are commonly used as backbone encoders by researchers \cite{zhang2020graph,hong2021weakly}, and we choose those two models as our encoders. 

\paragraph{Decoder.}
Tree-based solvers have been proven as an effective decoder in  \cite{xie2019goal,zhang2020graph,liang2021mwp}. The tree structure can decompose the goal and make the answer expression simpler. Empirically, we follow the decoder implementation of \cite{xie2019goal}. 

\subsection{Network Training}
Besides the backbone \textbf{MWP solver} (including encoder and decoder), our training framework has a \textbf{solution buffer} and a \textbf{solution discriminator}. 
The \textbf{buffer} stores the training data, including those initialized from the given training data, and those generated by an optimized \textbf{MWP solver}. The generated solutions for one given problem are different from the known solution to enlarge the training data diversity.   
To control the quality of buffered solutions to participate in training,  the \textbf{discriminator}   evaluates their qualification scores.  The solutions with higher qualification scores are allowed to contribute more in future training. 
The interactive process of the \textbf{MWP solver}, the  \textbf{buffer} and the \textbf{discriminator} follows three steps in an iterative manner. 
We introduce the details of the three steps in each iteration as follows.

\paragraph{Step 1: Probabilistic Training.}
In a standard fully supervised setting with training problem-equation ($W$-$S$) pairs, the MWP solver parameters of both encoder and decoder can be tuned by minimizing the following negative likelihood:
\begin{equation}
\theta^* = {\underset{\theta}{\text{argmin}}}{\{-\log P(S|W, \theta)\}},
    \label{eq:old}
\end{equation}
where $\theta^*$ covers all the parameters of the solver and is   the best to minimize the loss by finishing one entire training round at the present moment.
The MWP solver with  $\theta^*$  is thus supposed to recognize the correct solution $S$ for problem $W$, although its generalizability  may be limited due to the training experience with only a single $S$ for each problem $W$.

The buffer in our proposed framework is designed to include diverse alternative solutions for a problem $W$. Although initialized differently in different training settings,  the buffer $\mathcal{B}$ is updated during training to include more high-quality alternative solution equations. 
Note that at the beginning of the first Step 1 when an MWP solver with parameter $\theta$ has not been trained, the buffer $\mathcal{B}$ has only one equation of a problem $W$ or empty. In the fully supervised setting, $\mathcal{B}$ includes the  given single ground-truth $S$ for $W$. In the weakly supervised setting, $\mathcal{B}$ includes the generated  equation $S$ by WDA for $W$ if searchable, otherwise $\mathcal{B}$ is empty for $W$. In semi-weakly supervised setting, we only initialize the buffers of equation-annotated problems and leave others empty. Therefore, the quality score $a_i=1$  ($a_i=0$ when the buffer is empty) before starting to run Step 1 in the first iteration, even though at this moment $\theta$ is randomly initialized and the MWP solver performs badly on evaluating $P(B_i|W, \theta)$. 

After the first iteration, the buffer is updated (step 2) and the solutions in the buffer are assigned with their qualification scores $a_i$ (step 3). 
With these scored solutions  in buffer $\mathcal{B}$, the MWP solver parameter $\theta$ is optimized by a new objective function:
\begin{equation}
\theta^* = {\underset{\theta}{\text{argmin}}}{\{
    - \Sigma_{B_i\in \mathcal{B}} \;  a_i \log P(B_i|W, \theta) \}}.
    \label{eq:new}
\end{equation}
The quality weight $a_i$ enables an alternative solution $B_i$ to participate in the training process in a probabilistic manner, instead of only focusing on the single annotated ground truth. Initially, all available $B_i$ participate with the same probability ($a_i$=1). This solution quality weight $a_i$ will be updated in Step 3.  

\paragraph{Step 2: Solution Buffer Update.}
The trained MWP solver with optimized encoder and decoder $\theta^*$ in Step 1 is expected to generate the correct solutions for those training MWPs. 
To make full use of the well-trained solver,  we apply $k$-beam search to generate solutions with top-$k$ probabilities. 
\textcolor{black}{As long as the generated solutions lead to the correct value answer, we regard them as potential solutions.}
Then if the generated solutions do not exist in the current buffer $\mathcal{B}$, we update the buffer with the new solutions. In this way, the buffer $\mathcal{B}$ is filled by more diverse solutions. 

In the initial iterations, the solver $\theta^*$ may still have limited generalizability  since $\theta^*$ is optimized by using mostly (one-problem; one-solution) pairs. However, the top-$k$ generated solutions are the most likely reasonable alternative solutions since the solver $\theta^*$ does have a good understanding of the given problem. Even if some generated solutions are not good enough, our next step of qualification score evaluation can identify them and minimize their influence in the next update of  $\theta^*$.
After running   several iterations, the MWP solver has experienced diverse  solutions, improved its generalizability,  and then been able to generate even more diverse solutions.
This is an imitation of learning with self-correction in our human-like ways, i.e., gaining richer experience helps us to learn better. 

\paragraph{Step 3: Solutions Evaluation.}
There are two ways to calculate the solution quality weight  $a_i$ in Eq. (\ref{eq:new}). The first way is to rely on the MWP solver $\theta^*$, from which solutions are generated. Since the generated solutions selected in buffer $\mathcal{B}$ are those with high $P(B_i|W,\theta)$, their weights are further normalized in  $\mathcal{B}$ by
\begin{equation}
    s_i = \frac{P(B_i|W, \theta)}{\Sigma_{B_j\in \mathcal{B}}P(B_j|W, \theta)}.
    \label{eq:weight}
\end{equation}

Staying with the calculation of $a_i = s_i$ enlarges the buffer continuously with generated solutions. The diversity is boosted, however, the quality of generated solutions will drop.
To objectively evaluate the quality of one buffered solution $B_i$ for a problem $W$, we need a discriminator (classifier) to evaluate the fitness of $B_i$ and $W$, independent of the MWP solver $\theta^*$. Note that $s_i$ in Eq. (\ref{eq:weight}) is proportional to the probability of mapping the problem $W$ to a solution $B_i$. It only understands (by encoding) the problem $W$ and has no understanding of $B_i$.
Inspired by \cite{ijcai2021-485}, we build a solution discriminator by using a contrastive learning strategy to evaluate the qualification score of $B_i$ being a correct and diverse solution to $W$.


The discriminator encodes a solution $S$  to be   $Z_s$ by a bi-direction GRU, and encodes the problem $W$ to be $Z_w$ by the encoder of MWP solver. 
Then the score $t$ is calculated by the bilinear similarity between the mean vector of $Z_w$ and $Z_s$: ${t_{ws} =\sigma(\overline{Z_w}X_t\overline{Z_s})}$ where $X_t$ is a learnable matrix and \textcolor{black}{$\sigma$ is the Sigmoid function.}
The score $t_{ws}$ should be close to 1 if $S$ is one of  the correct solutions to $W$. Otherwise, $t_{ws}$ is 0. To train the discriminator  for this purpose, its parameter $\phi$ (covering $X_t$ and the bi-direction GRU of solution encoder) is optimized over the following objective function: 


\begin{equation}
 \begin{aligned}
\phi^* = {\underset{\phi}{\text{argmin}}} \{-\sum_{S \in \mathcal{S}^{pos}(W)} log (t_{ws}) \\ 
- \sum_{S \in \mathcal{S}^{neg}(W)}  log (1-t_{ws})  \} 
\end{aligned}
\label{loss_dis}
\end{equation}


The positive solutions $\mathcal{S}^{pos}(W)$ are generated by applying commutative law and associative law on the ground truth solution of $W$ to encourage the diversity of solutions.  In weakly supervised settings, we just use the solution generated by either WDA or the model as the positive equation, because there is no ground truth solution. 
The negative solutions $\mathcal{S}^{neg}(W)$ are generated by following  \cite{ijcai2021-485}, i.e., performing random manipulations on positive solutions with a disturbance probability $\lambda$. 
In this way,  the discriminator is trained to distinguish the diverse correct solutions and true negative solutions to problem $W$.
This is the first attempt to apply operation laws to encourage the diversity of solutions in solving  MWPs. We will show the effectiveness of this designed discriminator in the next section. Note that we do not perform positive sample augmentation in weakly supervised settings because there is no ground truth solution. We just use the solution generated by either WDA or the model as the positive equation. The details of the generation algorithm can be found in the next section.

\paragraph{Solution Augmentation.}
In order to augment the ground truth solutions to support the training of the discriminator. Let capital letters like $A, B, C$ denote a number or a complete equation that can lead to a number. For example, $A$ could be the number $2$, and also could be an equation $2+2$ that leads to $4$. We design the following rule-based solution augmentation and apply them on both full solutions and partial solutions (which still have to be equations):
\begin{itemize}
    \item For solutions with a shape $A + B$, we swap the positions $A$ and $B$ if there is no $*$ and $/$ connected to either $A$ or $B$.
    \item For solutions with a shape $A \pm B - C$ or $A - C \pm B$, we swap the positions of $\pm B$ and $-C$ if there is no $*$ and $/$ connected to either $B$ or $C$.
    \item For solutions with a shape $A * B$, we swap the positions of $A$ and $B$.
    \item For solutions with a shape $A * B / C$, $A / C * B$ or $A / B / C$, we swap the positions of $*B$ and $/C$ or the positions of $/B$ and $/C$.
\end{itemize}

\begin{algorithm}
\small
\caption{Iterative Training}
\label{alg:algorithm1}
\textbf{Input}: Problem $W$, Solution Buffer $S$, Solver $M$, Discriminator $D$  \\
\begin{algorithmic}[1] 
\FOR{Each Problem $W$ with a solution buffer $S$ in the dataset}
\IF{Epoch \textless 100}
\STATE Set $a_i$ in Eq. 2 as $s_i$ in Eq.3.
\ELSE
\STATE Set $a_i$ in Eq. 2 as $(s_i + t_{w{s_i}})/2$.
\ENDIF
\STATE Train the solver $M$ with Eq. 2 and buffer $S$.
\STATE Train the discriminator $D$ with Eq. 4.
\STATE {Apply beam-search on $M$ to solve $W$}
\STATE {Save equations that reach the correct value to buffer $S$}
\ENDFOR
\end{algorithmic}
\end{algorithm}
\paragraph{Training Process.}
Training the discriminator by Eq. (\ref{loss_dis}) needs a good MWP encoder to represent $W$ as $Z_w$. However, the problem encoder is updated with the decoder as a whole in the MWP solver by Eq. (\ref{eq:new}). In addition, the qualification score $t_{ws}$ from the discriminator is not usable to replace $a_i$ in  Eq. (\ref{eq:new}) until the discriminator is well trained. Therefore, to have a stable and effective training process, we organize the training epochs in two stages. In the first stage, e.g., the first 100 epochs, the weight $a_i$ in  Eq. (\ref{eq:new}) is set to $s_i$ in Eq.  (\ref{eq:weight}). The training process goes through step 1-3 iteratively: optimizing $\theta^*$, generating solutions to buffer, calculating $s_i$, and optimizing $\phi^*$. 
In the second stage when the discriminator is good enough to give reasonable scores, the weight $a_i$ is changed to $(s_i + t_{w{s_i}})/2$, taking into account the generation probability of the decoder and the fitness judged by the discriminator. Then the training process again goes through the iteration of optimizing $\theta^*$, generating solutions to buffer, calculating $a_i$ and optimizing $\phi^*$, until \textcolor{black}{the maximum epoch is reached.} For the sake of better understanding, we formulate our training process with a pseudo-code as shown in Alg. \ref{alg:algorithm1}.

\section{Experiments}
In this section, we first introduce the datasets and baselines that we use. Then we give a brief description about the computational environment and hyperparameters. For quantitative analysis, our solver outperforms all baseline methods under three different supervision in terms of accuracy. Besides, we conduct a solution diversity evaluation that shows our solver is able to generate multiple solutions in beam search, which confirms that the solution diversification ability is embedded into our solver. Also, we have an ablation study to show the contribution of different components. Due to the space limit, some qualitative experiments are located in our appendix.

\begin{table}
\renewcommand\arraystretch{1.2}
\centering
\begin{tabular}{|m{1.75cm}<{\centering}|m{5cm}<{\centering}|}
\hline
Problem Description: & Please calculate: 840/6/70+630 \\ 
\hline
Answer:       &  840/6/70+630 = 632    \\
\hline
\end{tabular}
\caption{An MWP example  having an explicit equation in the problem description, which are too easy, also not suitable for weak-supervised solver training and solution diversity evaluation.  
}
\label{tab:unweak}
\end{table}

\subsection{Used Datasets}
\paragraph{D-Math23k.}
Math23k contains 23,162 Chinese MWPs, which are annotated with equations as their solutions. 
In the Math23k dataset, there are some problems whose solutions are explicitly given in the problem description, as shown in Table \ref{tab:unweak}. It is more reasonable to discard those problems without solution diversity. After cleaning, there are 22,195 MWPs left in Math23k. And we call this subset D(iversity)-Math23k. We report the performance of 5-fold cross-validation on it following \cite{xie2019goal} and \cite{hong2021weakly}. Since we filtered out many simple problems and evaluate them under a more difficult setting, i.e. 5-fold cross-validation. Therefore, our re-produced performance on D-Math23k is not as good as some baseline \cite{zhang2020graph,liang2021mwp} reported.

\paragraph{Weak12k.}
We curate and release a novel math word problem (MWP) dataset called Weak12k with 12,117 MWPs. Each problem in this dataset is annotated with a final value answer instead of an equation solution. To our knowledge, Weak12k is the first Chinese MWP dataset in a weakly supervised manner. This dataset will be released to the public upon paper acceptance to facilitate future studies like semi-weakly supervised solver development. Compared with the most commonly used dataset Math23k, the problems in Weak12k are more difficult to solve, analysis and examples can be found in our appendix. 
In addition, the LBF method \cite{hong2021weakly} has a large performance gap on Math23k (over 50\%) and Weak12k (below 30\%) as shown in Table \ref{tab:result_semi}. Therefore, we believe that the new dataset Weak12k is  an indispensable  benchmark to the MWP community. In the experiment of this paper, we also report the results of 5-fold cross-validation on this dataset.

\begin{table}[t]
\renewcommand\arraystretch{1.2}
\centering
\setlength{\tabcolsep}{1.0mm}{
\begin{tabular}{|m{3cm}<{\centering}|m{2cm}<{\centering}|}
\hline
(Fully supervised)  & D-Math23k  \\
\hline
DNS         & $50.2$     \\
S-Aligned  & $55.4$       \\
GTS        & $65.7$   \\
Graph2Tree  & $66.6$ \\
MWP-BERT  & $69.2$ \\
\hline
D-GTS  & $67.1$ \\
D-Graph2Tree  & $68.9$ \\
D-MWP-BERT & $\mathbf{73.3}$ \\
\hline
\end{tabular}}
\caption{Comparison of answer accuracy (\%) on D-Math23k dataset under the fully supervised setting. The best results are in boldface. D-solver is our proposed diversity-injected solver.}
\label{tab:result_fully}
\end{table}
\subsection{Baselines}
In fully supervised setting, we select DNS \cite{wang2017deep}, S-Aligned \cite{DBLP:conf/naacl/Chiang2019}, GTS \cite{xie2019goal}, Graph2Tree\cite{zhang2020graph} and MWP-BERT \cite{liang2021mwp} as our baselines. DNS stands for the deep neural solver, which is a vanilla GRU-based Seq2Seq model. S-Aligned uses a stack to generate solutions. GTS presents a goal-driven tree-based solver. Graph2Tree develops a GNN-based encoder to capture more information about quantities and MWP-BERT develops a pre-trained-language-model-based encoder. For semi-weakly and weakly supervised settings, all other baselines like GTS, MWP-BERT cannot work. Therefore, we take \cite{hong2021weakly} as a baseline method, which uses a fixing mechanism to modify wrong solutions into correct ones for training.
\begin{table}
\renewcommand\arraystretch{1.1}
\centering
\setlength{\tabcolsep}{1.0mm}{
\begin{tabular}{|m{4.0cm}<{\centering}|m{1.75cm}<{\centering}|m{1.75cm}<{\centering}|}
\hline
& D-Math23k & Weak12k \\
\hline
\multicolumn{3}{|c|}{Semi-weakly supervised} \\
\hline
LBF         & $54.1$  & $33.6$     \\
\hline
D-GTS w WDA  & $67.1$  & $57.9$    \\
D-MWP-BERT w WDA  & $73.2$  & $67.5$    \\
D-GTS w/o WDA  & $68.4$  & $59.2$    \\
D-MWP-BERT w/o WDA  & $\mathbf{74.4}$  & $\mathbf{70.9}$    \\
\hline
\multicolumn{3}{|c|}{Weakly supervised}   \\
\hline
LBF         & $53.1$  & $29.5$     \\
\hline
D-GTS w WDA  & $55.2$  & $\mathbf{35.0}$    \\
D-MWP-BERT w WDA  & $\mathbf{56.0}$  & $31.8$    \\
\hline
\end{tabular}}
\caption{Comparison of answer accuracy (\%) on D-Math23k dataset under semi-weakly and weakly supervised setting. The best results are in boldface. ``w WDA'' indicates that we initialize the solution buffers for problems in Weak12k with the proposed WDA. ``w/o WDA'' means we leave the solution buffers for Weak12k empty.}
\label{tab:result_semi}
\end{table}
\subsection{Implementation Details}

We use Pytorch to construct the code and the NVIDIA RTX 2080Ti graphic card to train the solvers. The code and data can be found in \footnote{\url{https://github.com/LZhenwen/Solution_Diversity}}. The dimension of the embedding matrix is 128, and the dimension of all hidden features is 512. We train the model 200 epochs with the Adam optimizer \cite{kingma2014adam} and the learning rate 0.001, which will be halved every 30 epochs. For the first 100 epochs use $a_i=s_i$ and the remaining epochs use $(s_i + t_{w{s_i}})/2$. We update the solution buffer every 5 epochs of parameter learning, to leave sufficient time to train the model after one round of solution buffer updates. 

\subsection{Quantitative Results}
\paragraph{Fully Supervised Training.}
In this setting, we train the model with MWPs and ground truth equation solutions. Only the D-Math23k dataset is used because MWPs in Weak12k are not annotated with equations. Benefiting from our training method, the accuracies of GTS, Graph2Tree, and MWP-BERT are improved. The potential reason for such improvement is our training method enables the model to witness more diverse solutions which leads to a better generalization ability across MWPs.

\paragraph{Semi-weakly Supervised Training.}
In real-world scenarios, we usually get mixed data, some MWPs are annotated with equations and others are annotated with values. The baseline method LBF is not able to work well under this setting because 
it is specially designed for weakly supervised training only. We take a combination of D-Math23k and Weak12k for training and evaluate the answer accuracy on them separately for comparison. The results in Table \ref{tab:result_semi} show that our method outperforms baseline methods by a large margin. We also find that the proposed WDA is not necessary under the semi-weakly supervised setting, since the equations generated by WDA have lower quality than those generated by trained MWP solvers in the augmentation stage. Therefore, it is better to leave the solution buffers of Weak12k MWPs empty in the beginning and fill them with solutions generated by the model trained on D-Math23k. 

\begin{table}

\renewcommand\arraystretch{1.07}
\centering
\setlength{\tabcolsep}{2mm}{
\begin{tabular}{|c|c|c|c|}
\hline
 & Top-1 & Top-3 & Top-5 \\
\hline
\multicolumn{4}{|c|}{Fully supervised} \\
\hline
GTS         & $65.7$  & $38.3$ & $29.0$    \\
\hline
LBF         & $65.2$  & $59.3$ & $51.2$    \\
\hline
D-GTS      & $\mathbf{67.1}$  & $\mathbf{61.0}$ & $\mathbf{55.9}$     \\
\hline
\multicolumn{4}{|c|}{Weakly supervised}   \\
\hline
LBF         & $53.1$  & $47.1$  &$43.0$    \\
\hline
D-GTS w WDA  & $\mathbf{55.2}$  &$\mathbf{53.9}$ &$\mathbf{51.8}$   \\
\hline
\end{tabular}}
\caption{Comparison of top-$k$ answer accuracy (\%) on D-Math23k dataset. A higher accuracy on a larger $k$ indicates stronger generalizability to produce better and more diverse solutions.}
\label{tab:result_diversity}
\end{table}

\paragraph{Weakly Supervised Training.}
Our training method is also able to work under the weakly supervised setting with the help of WDA, when no equation-annotated problems are available. As we clarified in Section 3.2, WDA is necessary for weakly supervised setting because we always need equation-annotated problems to start training. Although WDA may generate spurious solutions, the model learns from them and achieves a satisfying performance. 
We evaluate on D-Math23k and Weak12k datasets in this setting separately. Experimental results in Table \ref{tab:result_semi} show that our proposed method is better than other baselines. The accuracy on Weak12k is lower than that on D-Math23k because the problems in Weak12k are generally more complicated and thus more difficult to solve.

\paragraph{Solution Diversity.}
Following LBF \cite{hong2021weakly}, we measure the overall answer accuracy in a complete beam search instead of only taking the first one. And we call this top-$k$ accuracy when the beam size is $k$. The higher accuracy of top-$k$ ($k > 1$) represents the more diverse solution that a solver can generate. The experiment results in Table \ref{tab:result_diversity} show that our solver outperforms GTS in the fully supervised setting and beats LBF in two different settings. We also conduct a qualitative analysis in our appendix, showing the effectiveness of our proposed solution diversification, by visualizing the Top-3 solutions in beam search. This diversity analysis strongly demonstrates the generalizability of solution diversity that our training method brings to the solver.

\begin{figure}
\centering 
\includegraphics[width=0.445\textwidth]{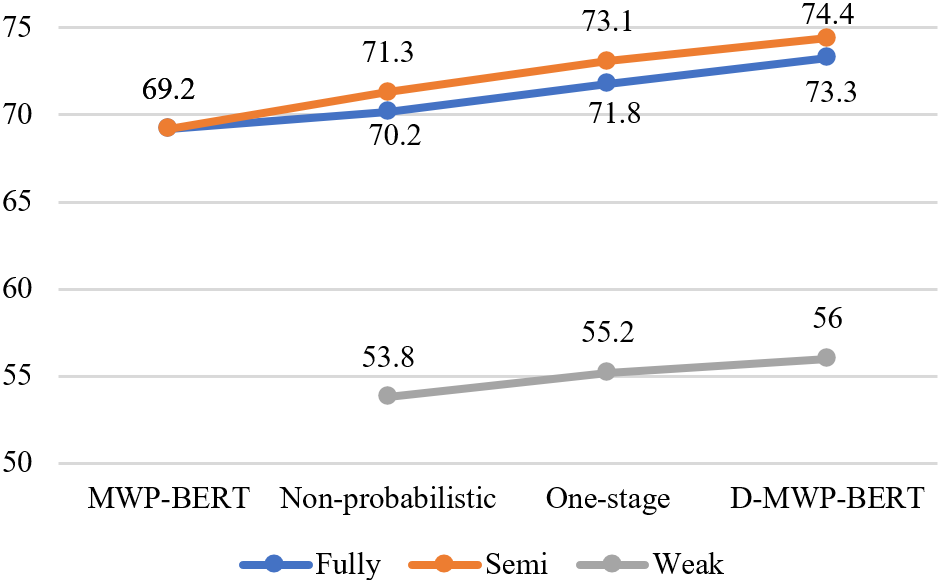} 
\caption{Ablation study under fully, semi-weakly and weakly supervised settings.
} 
\label{fig:ab} 
\end{figure}
\subsection{Ablation Study}
Some ablation on WDA is already included in Table \ref{tab:result_semi}. To further understand the effect of our two-stage training method, we conduct an analysis as shown in Figure \ref{fig:ab}. We use MWP-BERT as the backbone solver and conduct the experiment on the D-Math23k dataset across 3 different settings (Weak12k is also used in semi-weakly supervised setting). Non-probabilistic training means that we keep the solution buffer but assign the same weights on all solutions in the buffer, ignoring the quality differences among them. One-stage training is abandoning the qualifying stage and only using the model to score solutions. The result demonstrates that our approach is not only able to help the solver generate diverse solutions as the learning targets, but also diminish the bad effect of spurious solutions during the qualifying stage. 
%

\section{Conclusion}
We present a novel training framework, aiming to augment diverse solutions for MWPs and score them based on their quality. With the help of the solution buffer and the proposed buffer update method, our training framework is able to find multiple solutions for one MWP. A discriminator is trained by a contrastive learning mechanism to qualify solutions. Our method works under fully, semi-weakly, and weakly supervised situations to improve the accuracy of arbitrary MWP solvers. Moreover, we develop a simple but effective solution generation method called weak data augmentation (WDA) in the weakly supervised situation. 
\section*{Acknowledgement}
The research work is partially supported by the Internal Asia Research Collaboration Grant, University of Notre Dame. Thanks for all reviewers for their valuable comments.

\bibliography{aaai23}

\end{document}